\documentclass[10pt,twocolumn,letterpaper]{article}

\usepackage{wacv}
\usepackage{times}
\usepackage{epsfig}
\usepackage{graphicx}
\graphicspath{ {./figs/} }
\usepackage{amsmath}
\usepackage{amssymb}
\usepackage{bm}
\usepackage{algorithm}
\usepackage[noend]{algpseudocode}
\usepackage{booktabs}
\usepackage{multirow}
\usepackage{xcolor}
\usepackage[normalem]{ulem}

\def\eqref#1{equation~\ref{#1}}

\def\1{\bm{1}}
\newcommand{\train}{\mathcal{D_{\mathrm{train}}}}

\newcommand{\test}{\mathcal{D_{\mathrm{test}}}}

\def\vzero{{\bm{0}}}

\def\vmu{{\bm{\mu}}}
\def\vlambda{{\bm{\lambda}}}
\def\vtheta{{\bm{\theta}}}

\def\vc{{\bm{c}}}

\def\vu{{\bm{u}}}
\def\vv{{\bm{v}}}

\def\vx{{\bm{x}}}
\def\vy{{\bm{y}}}
\def\vz{{\bm{z}}}

\def\mI{{\bm{I}}}
\def\mJ{{\bm{J}}}

\def\mP{{\bm{P}}}

\def\mS{{\bm{S}}}

\def\mSigma{{\bm{\Sigma}}}

\DeclareMathAlphabet{\mathsfit}{\encodingdefault}{\sfdefault}{m}{sl}
\SetMathAlphabet{\mathsfit}{bold}{\encodingdefault}{\sfdefault}{bx}{n}

\def\sG{{\mathbb{G}}}

\DeclareMathOperator*{\argmin}{arg\,min}

\def\wacvPaperID{58} 

\wacvfinalcopy 

\ifwacvfinal
\def\assignedStartPage{1} 
\fi

\ifwacvfinal
\usepackage[breaklinks=true,bookmarks=false]{hyperref}
\else
\usepackage[pagebackref=true,breaklinks=true,colorlinks,bookmarks=false]{hyperref}
\fi

\ifwacvfinal
\else
\fi

\begin{document}

\title{CFLOW-AD: Real-Time Unsupervised Anomaly Detection with Localization\\ via Conditional Normalizing Flows}

\author{Denis~Gudovskiy\textsuperscript{\rm 1}
	\qquad~Shun~Ishizaka\textsuperscript{\rm 2}
	\qquad~Kazuki~Kozuka\textsuperscript{\rm 2}\\
	{\textsuperscript{\rm 1}Panasonic AI Lab, USA} ~~
	{\textsuperscript{\rm 2}Panasonic Technology Division, Japan} \\
	\small{\texttt{denis.gudovskiy@us.panasonic.com}}
	\qquad\small{\texttt{\{ishizaka.shun, kozuka.kazuki\}@jp.panasonic.com}}
}

\maketitle
\thispagestyle{empty}

\begin{abstract}
Unsupervised anomaly detection with localization has many practical applications when labeling is infeasible and, moreover, when anomaly examples are completely missing in the train data. While recently proposed models for such data setup achieve high accuracy metrics, their complexity is a limiting factor for real-time processing. In this paper, we propose a real-time model and analytically derive its relationship to prior methods. Our CFLOW-AD model is based on a conditional normalizing flow framework adopted for anomaly detection with localization. In particular, CFLOW-AD consists of a discriminatively pretrained encoder followed by a multi-scale generative decoders where the latter explicitly estimate likelihood of the encoded features. Our approach results in a computationally and memory-efficient model: CFLOW-AD is faster and smaller by a factor of 10$\times$ than prior state-of-the-art with the same input setting. Our experiments on the MVTec dataset show that CFLOW-AD outperforms previous methods by 0.36\% AUROC in detection task, by 1.12\% AUROC and 2.5\% AUPRO in localization task, respectively. We open-source our code with fully reproducible experiments\footnote{\href{https://github.com/gudovskiy/cflow-ad}{Our code is available at github.com/gudovskiy/cflow-ad}}.
\end{abstract}

\section{Introduction}
\label{sec:intro}
\begin{figure}[t]
	\centering
	\includegraphics[width=0.8\columnwidth]{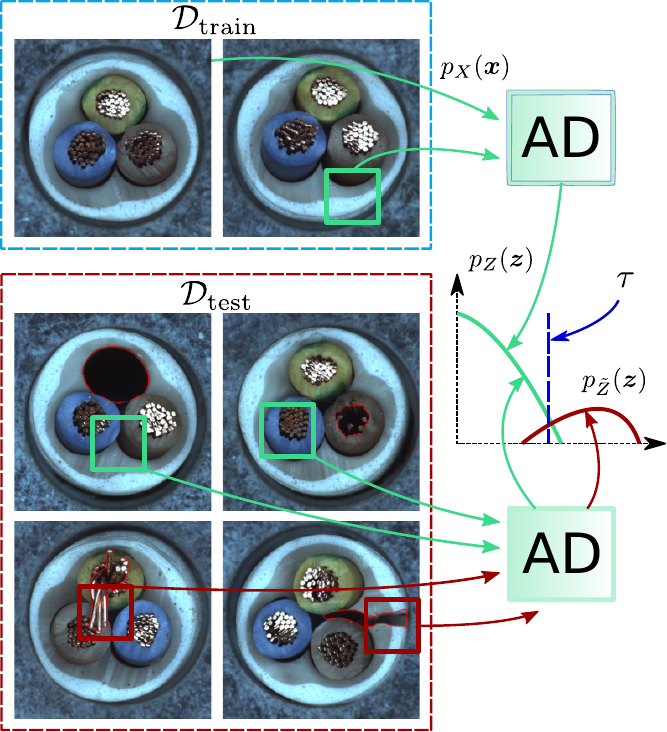}
	\caption{An example of the proposed out-of-distribution (OOD) detector for anomaly localization trained on anomaly-free $\train$ (top row). Sliced cable images are from the MVTec dataset~\cite{Bergmann_2019_CVPR}, where the bottom row illustrates $\test$ ground truth masks for anomalies (red) and the middle row shows examples of anomaly-free patches (green). The OOD detector learns the distribution of anomaly-free patches $\vx$ with $p_{X}(\vx)$ density and transforms it into a Gaussian distribution with $p_{Z}(\vz)$ density. Threshold $\tau$ separates in-distribution patches from the OOD patches with $p_{\tilde{Z}}(\vz)$ density.}
	\label{fig:problem}
\end{figure}

Anomaly detection with localization (AD) is a growing area of research in computer vision with many practical applications \eg industrial inspection~\cite{Bergmann_2019_CVPR}, road traffic monitoring~\cite{Li_2020_CVPR_Workshops}, medical diagnostics~\cite{zhou2020encoding} \etc. However, the common \textit{supervised} AD~\cite{Saleh13} is not viable in practical applications due to several reasons. First, it requires labeled data which is costly to obtain. Second, anomalies are usually \textit{rare long-tail examples} and have low probability to be acquired by sensors. Lastly, consistent labeling of anomalies is subjective and requires extensive domain expertise as illustrated in Figure~\ref{fig:problem} with industrial cable defects.

With these limitations of the supervised AD, a more appealing approach is to collect only \textit{unlabeled anomaly-free images} for train dataset $\train$ as in Figure~\ref{fig:problem} (top row). Then, any deviation from anomaly-free images is classified as an anomaly. Such data setup with low rate of anomalies is generally considered to be \textit{unsupervised}~\cite{Bergmann_2019_CVPR}. Hence, the AD task can be reformulated as a task of \textit{out-of-distribution detection} (OOD) with the AD objective.

While OOD for low-dimensional industrial sensors (\eg power-line or acoustic) can be accomplished using a common $k$-nearest-neighbor or more advanced clustering methods~\cite{knn-ad}, it is less trivial for high-resolution images. Recently, convolutional neural networks (CNNs) have gained popularity in extracting semantic information from images into downsampled feature maps~\cite{Bergmann_2019_CVPR}. Though feature extraction using CNNs has relatively low complexity, the post-processing of feature maps is far from real-time processing in the state-of-the-art unsupervised AD methods~\cite{defard2020padim}.

To address this complexity drawback, we propose a CFLOW-AD model that is based on conditional normalizing flows. CFLOW-AD is agnostic to feature map spatial dimensions similar to CNNs, which leads to a higher accuracy metrics as well as a lower computational and memory requirements. We present the main idea behind our approach in a toy OOD detector example in Figure~\ref{fig:problem}. A distribution of the anomaly-free image patches $\vx$ with probability density function $p_{X}(\vx)$ is learned by the AD model. Our translation-equivariant model is trained to transform the original distribution with $p_{X}(\vx)$ density into a Gaussian distribution with $p_{Z}(\vz)$ density. Finally, this model separates in-distribution patches $\vz$ with $p_{Z}(\vz)$ from the out-of-distribution patches with $p_{\tilde{Z}}(\vz)$ using a threshold $\tau$ computed as the Euclidean distance from the distribution mean.

\section{Related work}
\label{sec:related}
We review models\footnote{For comprehensive review of the existing AD methods we refer readers to Ruff~\etal~\cite{50044} and Pang~\etal~\cite{pang21} surveys.} that employ the data setup from Figure~\ref{fig:problem} and provide experimental results for popular MVTec dataset~\cite{Bergmann_2019_CVPR} with factory defects or Shanghai Tech Campus (STC) dataset~\cite{Luo_2017_ICCV} with surveillance camera videos. We highlight the research related to a more challenging task of \textit{pixel-level anomaly localization (segmentation)} rather than a more simple \textit{image-level anomaly detection}.

Napoletano~\etal~\cite{cnn_feature_dictionary_nanofibres} propose to use CNN feature extractors followed by a principal component analysis and $k$-mean clustering for AD. Their feature extractor is a ResNet-18~\cite{he} pretrained on a large-scale ImageNet dataset~\cite{alexnet}. Similarly, SPADE~\cite{cohen2021subimage} employs a Wide-ResNet-50~\cite{BMVC2016_87} with multi-scale pyramid pooling that is followed by a $k$-nearest-neighbor clustering. Unfortunately, clustering is slow at test-time with high-dimensional data. Thus, parallel convolutional methods are preferred in real-time systems.

Numerous methods are based on a natural idea of \textit{generative modeling}. Unlike models with the discriminatively-pretrained feature extractors~\cite{cnn_feature_dictionary_nanofibres, cohen2021subimage}, generative models learn distribution of anomaly-free data and, therefore, are able to estimate a proxy metrics for anomaly scores even for the unseen images with anomalies. Recent models employ generative adversarial networks (GANs)~\cite{schlegl2019_fast_anogan, schlegl_anogan} and variational autoencoders (VAEs)~\cite{c_baur_vae_gan, q_space_golkov}.

A fully-generative models~\cite{schlegl2019_fast_anogan, schlegl_anogan, c_baur_vae_gan, q_space_golkov} are directly applied to images in order to estimate pixel-level probability density and compute per-pixel reconstruction errors as anomaly scores proxies. These fully-generative models are unable to estimate the \textit{exact data likelihoods}~\cite{bergmann2018ssim, nalisnick2018do} and do not perform better than the traditional methods~\cite{cnn_feature_dictionary_nanofibres, cohen2021subimage} according to MVTec survey in~\cite{Bergmann_2019_CVPR}. Recent works~\cite{schirrmeister2020understanding, kirichenko2020normalizing} show that these models tend to capture only low-level correlations instead of relevant semantic information. To overcome the latter drawback, a hybrid DFR model~\cite{DFR2020} uses a pretrained feature extractor with multi-scale pyramid pooling followed by a convolutional autoencoder (CAE). However, DFR model is unable to estimate the exact likelihoods.

Another line of research proposes to employ a student-teacher type of framework~\cite{Bergmann_2020_CVPR, salehi2020multiresolution, wang2021studentteacher}. Teacher is a pretrained feature extractor and student is trained to estimate a scoring function for AD. Unfortunately, such frameworks underperform compared to state-of-the-art models.

Patch SVDD~\cite{Yi_2020_ACCV} and CutPaste~\cite{li2021cutpaste} introduce a self-supervised pretraining scheme for AD. Moreover, Patch SVDD proposes a novel method to combine multi-scale scoring masks to a final anomaly map. Unlike the nearest-neighbor search in~\cite{Yi_2020_ACCV}, CutPaste estimates anomaly scores using an efficient Gaussian density estimator. While the self-supervised pretraining can be helpful in uncommon data domains, Schirrmeister~\etal~\cite{schirrmeister2020understanding} argue that large natural-image datasets such as ImageNet can be a more representative for pretraining compared to a small application-specific datasets \eg industrial MVTec~\cite{Bergmann_2019_CVPR}.

The state-of-the-art PaDiM~\cite{defard2020padim} proposes surprisingly simple yet effective approach for anomaly localization. Similarly to~\cite{DFR2020, cohen2021subimage, Yi_2020_ACCV}, this approach relies on ImageNet-pretrained feature extractor with multi-scale pyramid pooling. However, instead of slow test-time clustering in~\cite{cohen2021subimage} or nearest-neighbor search in~\cite{Yi_2020_ACCV}, PaDiM uses a well-known Mahalanobis distance metric~\cite{mahalanobis1936generalized} as an anomaly score. The metric parameters are estimated for each feature vector from the pooled feature maps. PaDiM has been inspired by Rippel~\etal~\cite{rippel2020modeling} who firstly advocated to use this measure for anomaly detection without localization.

DifferNet~\cite{rudolph2020differnet} uses a promising class of generative models called \textit{normalizing flows} (NFLOWs)~\cite{45819} for image-level AD. The main advantage of NFLOW models is ability to estimate the exact likelihoods for OOD compared to other generative models~\cite{schlegl2019_fast_anogan, schlegl_anogan, c_baur_vae_gan, q_space_golkov, DFR2020}. In this paper, we extend DifferNet approach to pixel-level anomaly localization task using our CFLOW-AD model. In contrast to RealNVP~\cite{45819} architecture with global average pooling in~\cite{rudolph2020differnet}, we propose to use conditional normalizing flows~\cite{ardizzone2019guided} to make CFLOW-AD suitable for low-complexity processing of multi-scale feature maps for localization task. We develop our CFLOW-AD with the following contributions:
\begin{itemize}
	\itemsep0em
	\item Our theoretical analysis shows why multivariate Gaussian assumption is a justified prior in previous models and why a more general NFLOW framework objective converges to similar results with the less compute.
	\item We propose to use conditional normalizing flows for unsupervised anomaly detection with localization using computational and memory-efficient architecture.
	\item We show that our model outperforms previous state-of-the art in both detection and localization due to the unique properties of the proposed CFLOW-AD model.
\end{itemize}

\begin{figure*}[t]
	\centering
	\includegraphics[width=0.9\textwidth]{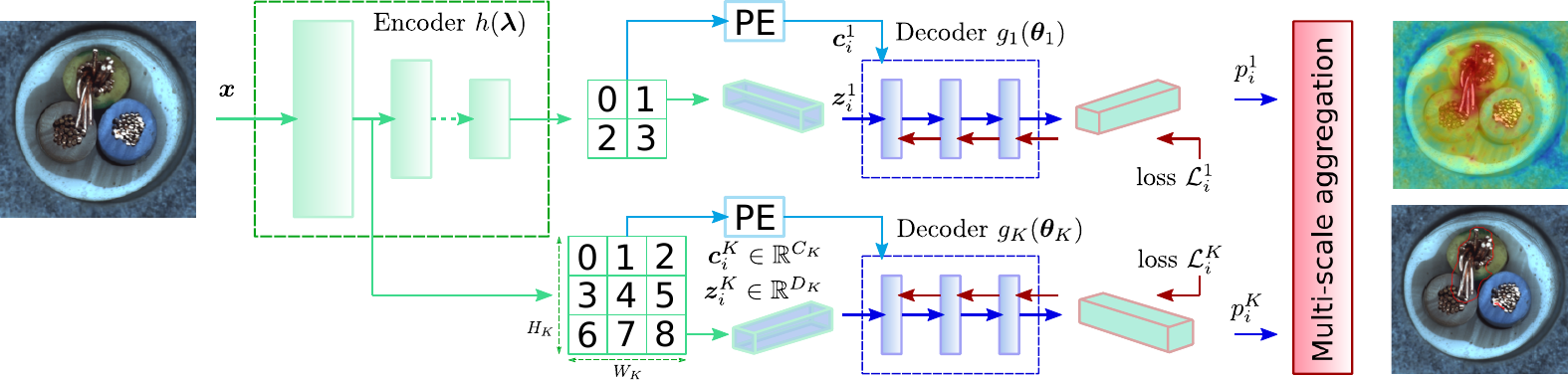}
	\caption{Overview of our CFLOW-AD with a fully-convolutional translation-equivariant architecture. Encoder $h(\vlambda)$ is a CNN feature extractor with multi-scale pyramid pooling. Pyramid pooling captures both global and local semantic information with the growing from top to bottom receptive fields. Pooled feature vectors $\vz^k_i$ are processed by a set of decoders $g_k(\vtheta_k)$ independently for each $k$th scale. Our decoder is a conditional normalizing flow network with a feature input $\vz^k_i$ and a conditional input $\vc^k_i$ with spatial information from a positional encoder (PE). The estimated multi-scale likelihoods $p^k_i$ are upsampled to the input size and added up to produce anomaly map.}
	\label{fig:cflow}
\end{figure*}

\section{Theoretical background}
\label{sec:theory}

\subsection{Feature extraction with Gaussian prior}
\label{sec:gauss}
Consider a CNN $h(\vlambda)$ trained for classification task. Its parameters $\vlambda$ are usually found by minimizing Kullback-Leibler ($D_{KL}$) divergence between joint train data distribution $Q_{\vx,\vy}$ and the learned model distribution $P_{\vx,\vy}(\vlambda)$, where $(\vx,\vy)$ is an input-label pair for supervised learning.

Typically, the parameters $\vlambda$ are initialized by the values sampled from the Gaussian distribution~\cite{He_2015_ICCV} and optimization process is regularized as
\begin{equation} \label{eq:th1}
\argmin_{\vlambda} D_{KL} \left[ Q_{\vx,\vy} \| P_{\vx,\vy}(\vlambda) \right] + \alpha R(\vlambda),
\end{equation}
where $R(\vlambda)$ is a regularization term and $\alpha$ is a hyperparameter that defines regularization strength.

The most popular CNNs~\cite{he, BMVC2016_87} are trained with $L_2$ weight decay~\cite{NIPS1991_8eefcfdf} regularization ($R(\vlambda) = \| \vlambda \|_2^2$). That imposes multivariate Gaussian (MVG) prior not only to parameters $\vlambda$, but also to the feature vectors $\vz$ extracted from the feature maps of $h(\vlambda)$~\cite{GoodBengCour16} intermediate layers.

\subsection{A case for Mahalanobis distance}
\label{sec:mahal}
With the same MVG prior assumption, Lee~\etal~\cite{NEURIPS2018_abdeb6f5} recently proposed to model distribution of feature vectors $\vz$ by MVG density function and to use Mahalanobis distance~\cite{mahalanobis1936generalized} as a confidence score in CNN classifiers. Inspired by~\cite{NEURIPS2018_abdeb6f5}, Rippel~\etal~\cite{rippel2020modeling} adopt Mahalanobis distance for anomaly detection task since this measure determines a distance of a particular feature vector $\vz$ to its MVG distribution. Consider a MVG distribution $\mathcal{N}(\vmu, \mSigma)$ with a density function $p_{Z}(\vz)$ for random variable $\vz \in \mathbb{R}^D$ defined as
\begin{equation} \label{eq:th2}
p_{Z}(\vz) = (2\pi)^{-D/2} \det \mSigma^{-1/2}  e^{-\frac{1}{2} (\vz - \vmu)^T \mSigma^{-1} (\vz - \vmu)},
\end{equation}
where $\vmu \in \mathbb{R}^D$ is a mean vector and $\mSigma \in \mathbb{R}^{D \times D}$ is a covariance matrix of a true anomaly-free density $p_{Z}(\vz)$.

Then, the Mahalanobis distance $M(\vz)$ is calculated as
\begin{equation} \label{eq:th3}
M(\vz) = \sqrt{ (\vz - \vmu)^T \mSigma^{-1} (\vz - \vmu)},
\end{equation}

Since the true anomaly-free data distribution is unknown, mean vector and covariance matrix from (\ref{eq:th3}) are replaced by the estimates $\hat{\vmu}$ and $\hat{\mSigma}$ calculated from the empirical train dataset $\train$. At the same time, density function $p_{\tilde{Z}}(\vz)$ of anomaly data has different $\tilde{\vmu}$ and $\tilde{\mSigma}$ statistics, which allows to separate out-of-distribution and in-distribution feature vectors using $M(\vz)$ from (\ref{eq:th3}).

This framework with MVG distribution assumption shows its effectiveness in image-level anomaly detection task~\cite{rippel2020modeling} and is adopted by the state-of-the-art PaDiM~\cite{defard2020padim} model in pixel-level anomaly localization task.

\subsection{Relationship with the flow framework}
\label{sec:nflow}
Dinh~\etal~\cite{45819} introduce a class of generative probabilistic models called normalizing flows. These models apply change of variable formula to fit an arbitrary density $p_{Z}(\vz)$ by a tractable base distribution with $p_{U}(\vu)$ density and a bijective invertible mapping $g^{-1}: Z \rightarrow U$. Then, the $\log$-likelihood of any $\vz \in Z$ can be estimated by
\begin{equation} \label{eq:th4}
\log \hat{p}_{Z} (\vz, \vtheta) = \log p_{U}(\vu) + \log \left| \det \mJ \right|,
\end{equation}
where a sample $\vu \sim p_{U}$ is usually from standard MVG distribution $(\vu \sim \mathcal{N}(\vzero, \mI))$ and a matrix $\mJ = \nabla_{\vz} g^{-1}(\vz, \vtheta)$ is the Jacobian of a bijective invertible flow model $(\vz = g(\vu, \vtheta)$ and $\vu = g^{-1}(\vz, \vtheta))$ parameterized by vector $\vtheta$.

The flow model $g(\vtheta)$ is a set of basic layered transformations with tractable Jacobian determinants. For example, $\log | \det \mJ |$ in RealNVP~\cite{45819} coupling layers is a simple sum of layer's diagonal elements. These models are optimized using stochastic gradient descent by maximizing $\log$-likelihood in~(\ref{eq:th4}). Equivalently, optimization can be done by minimizing the reverse $D_{KL} \left[ \hat{p}_{Z} (\vz, \vtheta) \| p_{Z} (\vz) \right]$~\cite{JMLR:v22:19-1028}, where $\hat{p}_{Z} (\vz, \vtheta)$ is the model prediction and $p_{Z} (\vz)$ is a target density. The loss function for this objective is defined as
\begin{equation} \label{eq:th5}
\mathcal{L}(\vtheta) = \mathbb{E}_{\hat{p}_{Z} (\vz, \vtheta)} \left[ \log \hat{p}_{Z} (\vz, \vtheta) - \log p_{Z} (\vz) \right].
\end{equation}

If $p_{Z} (\vz)$ is distributed according to Section~\ref{sec:gauss} MVG assumption, we can express~(\ref{eq:th5}) as a function of Mahalanobis distance $M(\vz)$ using its definition from (\ref{eq:th3}) as
\begin{equation} \label{eq:th6}
\mathcal{L}(\vtheta) = \mathbb{E}_{\hat{p}_{Z} (\vz, \vtheta)} \left[ \frac{M^{2}(\vz) - E^{2}(\vu)}{2} + \log \frac{| \det \mJ |}{ \det \mSigma^{-1/2}} \right],
\end{equation}
where $E^2(\vu) = \| \vu \|_2^2$ is a squared Euclidean distance of a sample $\vu \sim \mathcal{N}(\vzero, \mI)$ (detailed proof in Appendix~A).

Then, the loss in (\ref{eq:th6}) converges to zero when the likelihood contribution term $| \det \mJ |$ of the model $g(\vtheta)$ (normalized by $\det \mSigma^{-1/2}$) compensates the difference between a squared Mahalanobis distance for $\vz$ from the target density and a squared Euclidean distance for $\vu \sim \mathcal{N}(\vzero, \mI)$.

This normalizing flow framework can estimate the exact likelihoods of any arbitrary distribution with $p_{Z}$ density, while Mahalanobis distance is limited to MVG distribution only. For example, CNNs trained with $L_1$ regularization would have Laplace prior~\cite{GoodBengCour16} or have no particular prior in the absence of regularization. Moreover, we introduce conditional normalizing flows in the next section and show that they are more compact in size and have fully-convolutional parallel architecture compared to~\cite{cohen2021subimage, defard2020padim} models.

\section{The proposed CFLOW-AD model}
\label{sec:our_nf}
\subsection{CFLOW encoder for feature extraction}
\label{sec:our_encoder}
We implement a feature extraction scheme with multi-scale feature pyramid pooling similar to recent models~\cite{cohen2021subimage, defard2020padim}. We define the discriminatively-trained CNN feature extractor as an encoder $h(\vlambda)$ in Figure~\ref{fig:cflow}. The CNN encoder maps image patches $\vx$ into a feature vectors $\vz$ that contain relevant semantic information about their content. CNNs accomplish this task efficiently due to their translation-equivariant architecture with the shared kernel parameters. In our experiments, we use ImageNet-pretrained encoder following Schirrmeister~\etal~\cite{schirrmeister2020understanding} who show that large natural-image datasets can serve as a representative distribution for pretraining. If a large application-domain unlabeled data is available, the self-supervised pretraining from~\cite{Yi_2020_ACCV, li2021cutpaste} can be a viable option.

One important aspect of a CNN encoder is its \textit{effective receptive field}~\cite{NIPS2016_c8067ad1}. Since the effective receptive field is not strictly bounded, the size of encoded patches $\vx$ cannot be exactly defined. At the same time, anomalies have various sizes and shapes, and, ideally, they have to be processed with the variable receptive fields. To address the ambiguity between CNN receptive fields and anomaly variability, we adopt common multi-scale feature pyramid pooling approach. Figure~\ref{fig:cflow} shows that the feature vectors $\vz^k_i \in\mathbb{R}^{D_k}, i \in \{ H_k \times W_k \}$ are extracted by $K$ pooling layers. Pyramid pooling captures both local and global patch information with small and large receptive fields in the first and last CNN layers, respectively. For convenience, we number pooling layers in the last to first layer order.

\subsection{CFLOW decoders for likelihood estimation}
\label{sec:our_decoder}
We use the general normalizing flow framework from Section~\ref{sec:nflow} to estimate $\log$-likelihoods of feature vectors $\vz$. Hence, our generative decoder model $g(\vtheta)$ aims to fit true density $p_Z(\vz)$ by an estimated parameterized density $\hat{p}_{Z} (\vz, \vtheta)$ from~(\ref{eq:th1}). However, the feature vectors are assumed to be independent of their spatial location in the general framework. To increase efficacy of distribution modeling, we propose to incorporate spatial prior into $g(\vtheta)$ model using conditional flow framework. In addition, we model $\hat{p}^k_{Z} (\vz, \vtheta)$ densities using $K$ independent decoder models $g_k(\vtheta_k)$ due to multi-scale feature pyramid pooling setup.

Our conditional normalizing flow (CFLOW) decoder architecture is presented in Figure~\ref{fig:cflow}. We generate a conditional vector $\vc^k_i$ using a 2D form of conventional positional encoding (PE)~\cite{NIPS2017_3f5ee243}. Each $\vc^k_i \in\mathbb{R}^{C_k}$ contains $\sin$ and $\cos$ harmonics that are unique to its spatial location $(h_k, w_k)_i$. We extend unconditional flow framework to CFLOW by concatenating the intermediate vectors inside decoder coupling layers with the conditional vectors $\vc_i$ as in~\cite{ardizzone2019guided}.

Then, the $k$th CFLOW decoder contains a sequence of conventional coupling layers with the additional conditional input. Each coupling layer comprises of fully-connected layer with $(D_k+C_k) \times (D_k+C_k)$ kernel, softplus activation and output vector permutations. Usually, the conditional extension does not increase model size since $C_k \ll D_k$. For example, we use the fixed $C_k = 128$ in all our experiments. Our CFLOW decoder has translation-equivariant architecture, because it slides along feature vectors extracted from the intermediate feature maps with kernel parameter sharing. As a result, both the encoder $h(\vlambda)$ and decoders $g_k(\vtheta_k)$ have convolutional translation-equivariant architectures.

We train CFLOW-AD using a maximum likelihood objective, which is equivalent to minimizing loss defined by
\begin{equation} \label{eq:cflow1}
\begin{split}
\mathcal{L}(\vtheta) &= D_{KL} \left[ p_{Z} (\vz) \right \| \hat{p}_{Z} (\vz, \vc, \vtheta)] \approx \\
&\frac{1}{N} \sum^N_{i=1} \left[ \frac{\| \vu_i \|_2^2}{2} - \log \left| \det \mJ_i \right| \right] + \text{const},
\end{split}
\end{equation}
where the random variable $\vu_i = g^{-1}(\vz_i, \vc_i, \vtheta)$, the Jacobian $\mJ_i = \nabla_{\vz} g^{-1}(\vz_i, \vc_i, \vtheta)$ for CFLOW decoder and an expectation operation in $D_{KL}$ is replaced by an empirical train dataset $\train$ of size $N$. For brevity, we drop the $k$th scale notation. The derivation is given in Appendix~B.

After training the decoders $g_k(\vtheta_k)$ for all $K$ scales using (\ref{eq:cflow1}), we estimate test dataset $\test$ $\log$-likelihoods as 
\begin{equation} \label{eq:cflow2}
\log \hat{p}_Z (\vz_i, \vc_i, \hat{\vtheta}) = - \frac{\| \vu_i \|_2^2 + D \log (2\pi)}{2} + \log \left| \det \mJ_i \right|.
\end{equation}

Next, we convert $\log$-likelihoods to probabilities $p^k_i = e^{\log \hat{p}_Z (\vz^k_i, \vc^k_i, \hat{\vtheta}_k)}$ for each $k$th scale using (\ref{eq:cflow2}) and normalize them to be in $\left[0:1\right]$ range. Then, we upsample $p^k_i$ to the input image resolution ($H \times W$) using bilinear interpolation $\mP_k =  b(p^k) \in \mathbb{R}^{H \times W} $. Finally, we calculate anomaly score maps $\mS^{H \times W}$ by aggregating all upsampled probabilities as $\mS = \max_{\sum^K_{k=1} \mP_k} - \sum^K_{k=1} \mP_k$.

\begin{table}[t]
	\caption{Complexity estimates for SPADE~\cite{cohen2021subimage}, PaDiM~\cite{defard2020padim} and our CFLOW-AD. We compare train and test complexity as well as memory requirements. All models use the same encoder $h(\vlambda)$ setup, but diverge in the post-processing. SPADE allocates memory for a train gallery $\sG$ used in $k$-nearest-neighbors. PaDiM keeps large matrices $(\mSigma^{k}_{i})^{-1}, i \in \{ H_k \times W_k \}$ for Mahalanobis distance. Our model employs trained decoders $g_k(\vtheta_k)$ for post-processing.}
	\label{tab:comp_est}
	\centering
	\begin{tabular}{cccc}
		\toprule
		Model   & Train & Test & Memory \\
		\midrule
		SPADE~\cite{cohen2021subimage}& $\sG$ & $\sum\nolimits_{\vv \in \sG} \|\vv-\vz_i\|_2^2$ & $\vlambda+\sG$ \\
		PaDiM~\cite{defard2020padim}  & $\mSigma^{-1}_{i}$ & $M(\vz_i)$ & $\vlambda+\mSigma^{-1}_{i}$ \\
		Ours& $\mathcal{L}(\vtheta)$  & $ \log \hat{p}_Z (\vz_i, \vc_i, \vtheta)$ & $\vlambda+\vtheta$ \\
		\bottomrule
	\end{tabular}
\end{table}

\subsection{Complexity analysis}
\label{sec:our_complexity}
Table~\ref{tab:comp_est} analytically compares complexity of CFLOW-AD and recent state-of-the-art models with the same pyramid pooling setup \ie $\vz^{k \in \{1 \ldots K\}}=h(\vx, \vlambda)$.

SPADE~\cite{cohen2021subimage} performs $k$-nearest-neighbor clustering between each test point $\vz^k_i$ and a gallery $\sG$ of train data. Therefore, the method requires large memory allocation for gallery $\sG$ and a clustering procedure that is typically slow compared to convolutional methods.

PaDiM~\cite{defard2020padim} estimates train-time statistics \ie inverses of covariance matrices $(\mSigma^{k}_{i})^{-1} \in\mathbb{R}^{D_k \times D_k}, i \in \{ H_k \times W_k \}$ to calculate  $M(\vz^k_i)$ at test-time. Hence, it has low computational complexity, but it stores in memory $H_k \times W_k$ matrices of $D_k \times D_k$ size for every $k$th pooling layer.

Our method optimizes generative decoders $g_k(\vtheta_k)$ using~(\ref{eq:cflow1}) during the train phase. At the test phase, CFLOW-AD simply infers data $\log$-likelihoods $\log \hat{p}_Z (\vz_i, \vc_i, \hat{\vtheta})$ using~(\ref{eq:cflow2}) in a fully-convolutional fashion. Decoder parameters $\vtheta_{k \in \{1 \ldots K\}}$ are relatively small as reported in Table~\ref{tab:comp_eval}.

\section{Experiments}
\label{sec:eval}

\begin{table*}[ht]
	\caption{Ablation study of CFLOW-AD using localization AUROC metric on the MVTec~\cite{Bergmann_2019_CVPR} dataset, \%. We experiment with input image resolution ($H \times W$), encoder architecture (ResNet-18 (R18), WideResnet-50 (WRN50) and MobileNetV3L (MNetV3)), type of normalizing flow (unconditional (UFLOW) and conditional (CFLOW)), number of coupling (\# of CL) and pooling layers (\# of PL).}
	\label{tab:ablation-results}
	\centering
	\begin{tabular}{c|ccccc|cc|cc}
		\toprule
		Encoder & WRN50 & WRN50 & WRN50 & WRN50 & WRN50 & R18 & R18 & MNetV3 & MNetV3 \\
		\midrule
		\# of CL & \multicolumn{2}{c}{4 $\quad \longrightarrow \quad$ 8} & 8 & 8 & 8 & 8 & 8 & 8 & 8\\
		\midrule
		\# of PL & 2 & \multicolumn{2}{c}{2 $\quad \longrightarrow \quad$ 3} & 3 & 3 & 3 & 3 & 3 & 3\\
		\midrule		
		H$\times$W & 256 & 256 & \multicolumn{2}{c}{256 $\quad \longrightarrow \quad$ 512} & 512 & \multicolumn{2}{c}{256 $\quad \longrightarrow \quad$ 512} & \multicolumn{2}{c}{256 $\quad \longrightarrow \quad$ 512}\\
		\midrule
		Type & CFLOW & CFLOW & CFLOW & \multicolumn{2}{c|}{CFLOW $\rightarrow$ UFLOW} & CFLOW & CFLOW & CFLOW & CFLOW\\
		\midrule
		Bottle     & 97.28\tiny$\pm$0.03 & 97.24\tiny$\pm$0.03 & 98.76\tiny$\pm$0.01 & \textbf{98.98}\tiny$\pm$0.01 & 98.83\tiny$\pm$0.01 & 98.47\tiny$\pm$0.03 & \textbf{98.64}\tiny$\pm$0.01 & 98.74 & \textbf{98.92} \\
		Cable      & 95.71\tiny$\pm$0.01 & 96.17\tiny$\pm$0.07 & \textbf{97.64}\tiny$\pm$0.04 & 97.12\tiny$\pm$0.06 & 95.29\tiny$\pm$0.04 & \textbf{96.75}\tiny$\pm$0.04 & 96.07\tiny$\pm$0.06 & \textbf{97.62} & 97.49 \\
		Capsule    & 98.17\tiny$\pm$0.02 & 98.19\tiny$\pm$0.05 & \textbf{98.98}\tiny$\pm$0.00 & 98.64\tiny$\pm$0.02 & 98.40\tiny$\pm$0.12 & \textbf{98.62}\tiny$\pm$0.02 & 98.28\tiny$\pm$0.05 & \textbf{98.89} & 98.75 \\
		Carpet     & 98.50\tiny$\pm$0.01 & 98.55\tiny$\pm$0.01 & 99.23\tiny$\pm$0.01 & \textbf{99.25}\tiny$\pm$0.01 & 99.24\tiny$\pm$0.00 & 99.00\tiny$\pm$0.01 & \textbf{99.29}\tiny$\pm$0.00 & 98.64 & \textbf{99.00} \\
		Grid       & 93.77\tiny$\pm$0.05 & 93.88\tiny$\pm$0.16 & 96.89\tiny$\pm$0.02 & \textbf{98.99}\tiny$\pm$0.02 & 98.74\tiny$\pm$0.00 & 93.95\tiny$\pm$0.04 & \textbf{98.53}\tiny$\pm$0.01 & 94.75 & \textbf{98.81} \\
		Hazelnut   & 98.08\tiny$\pm$0.01 & 98.13\tiny$\pm$0.02 & 98.82\tiny$\pm$0.01 & \textbf{98.89}\tiny$\pm$0.01 & 98.88\tiny$\pm$0.01 & \textbf{98.81}\tiny$\pm$0.01 & 98.41\tiny$\pm$0.01 & 98.88 & \textbf{99.00} \\
		Leather    & 98.92\tiny$\pm$0.02 & 99.00\tiny$\pm$0.06 & 99.61\tiny$\pm$0.01 & \textbf{99.66}\tiny$\pm$0.00 & 99.65\tiny$\pm$0.00 & 99.45\tiny$\pm$0.01 & \textbf{99.51}\tiny$\pm$0.02 & 99.50 & \textbf{99.64} \\
		Metal Nut  & 96.72\tiny$\pm$0.03 & 96.72\tiny$\pm$0.06 & \textbf{98.56}\tiny$\pm$0.03 & 98.25\tiny$\pm$0.04 & 98.16\tiny$\pm$0.03 & \textbf{97.59}\tiny$\pm$0.05 & 96.42\tiny$\pm$0.03 & 98.36 & \textbf{98.78} \\
		Pill       & 98.46\tiny$\pm$0.02 & 98.46\tiny$\pm$0.01 & \textbf{98.95}\tiny$\pm$0.00 & 98.52\tiny$\pm$0.05 & 98.20\tiny$\pm$0.08 & \textbf{98.34}\tiny$\pm$0.02 & 97.80\tiny$\pm$0.05 & \textbf{98.69} & 98.44 \\
		Screw      & 94.98\tiny$\pm$0.06 & 95.28\tiny$\pm$0.06 & 98.10\tiny$\pm$0.05 & \textbf{98.86}\tiny$\pm$0.02 & 98.78\tiny$\pm$0.01 & 97.38\tiny$\pm$0.03 & \textbf{98.40}\tiny$\pm$0.03 & 98.04 & \textbf{99.09} \\
		Tile       & 95.52\tiny$\pm$0.02 & 95.66\tiny$\pm$0.06 & 97.71\tiny$\pm$0.02 & \textbf{98.01}\tiny$\pm$0.01 & 97.98\tiny$\pm$0.02 & 95.10\tiny$\pm$0.02 & \textbf{95.80}\tiny$\pm$0.10 & 96.07 & \textbf{96.48} \\
		Toothbrush & 98.02\tiny$\pm$0.03 & 97.98\tiny$\pm$0.00 & 98.56\tiny$\pm$0.02 & \textbf{98.93}\tiny$\pm$0.00 & 98.89\tiny$\pm$0.00 & 98.44\tiny$\pm$0.02 & \textbf{99.00}\tiny$\pm$0.01 & 98.09 & \textbf{98.80} \\
		Transistor & 93.09\tiny$\pm$0.28 & 94.05\tiny$\pm$0.11 & \textbf{93.28}\tiny$\pm$0.40 & 80.52\tiny$\pm$0.13 & 76.28\tiny$\pm$0.14 & \textbf{92.71}\tiny$\pm$0.23 & 83.34\tiny$\pm$0.46 & \textbf{97.79} & 95.22 \\
		Wood       & 90.65\tiny$\pm$0.10 & 90.59\tiny$\pm$0.07 & 94.49\tiny$\pm$0.03 & \textbf{96.65}\tiny$\pm$0.01 & 96.56\tiny$\pm$0.02 & 93.51\tiny$\pm$0.03 & \textbf{95.00}\tiny$\pm$0.04 & 92.24 & \textbf{94.96} \\
		Zipper     & 96.80\tiny$\pm$0.02 & 97.01\tiny$\pm$0.05 & 98.41\tiny$\pm$0.09 & \textbf{99.08}\tiny$\pm$0.02 & 99.06\tiny$\pm$0.01 & 97.71\tiny$\pm$0.06 & \textbf{98.98}\tiny$\pm$0.01 & 97.50 & \textbf{99.07} \\
		\midrule
		Average    & 96.31               & 96.46               &      97.87 &                        97.36 &               96.86 &               97.06 &              96.90  & 97.59 & 98.16 \\
		\bottomrule
	\end{tabular}
\end{table*}

\begin{table*}[ht]
	\caption{The detailed comparison of PaDiM~\cite{defard2020padim}, SPADE~\cite{cohen2021subimage}, CutPaste~\cite{li2021cutpaste} and our CFLOW-AD on the MVTec~\cite{Bergmann_2019_CVPR} dataset for every class using AUROC or, if available, a tuple (AUROC, AUPRO) metric, \%. CFLOW-AD model is with the best hyperparameters from Section~\ref{subsec:ablation_eval} ablation study. For fair comparison, we group together results with the same encoder architectures such as ResNet-18 and WideResNet-50.}
	\label{tab:class-results}
	\centering
	\begin{tabular}{c|cc|cccc|ccc}
		\toprule
		Task & \multicolumn{2}{|c}{Localization} & \multicolumn{4}{|c}{Detection} & \multicolumn{3}{|c}{Localization} \\
		\midrule
		Encoder & \multicolumn{4}{|c|}{ResNet-18} & EffNetB4 & \multicolumn{4}{|c}{WideResNet-50} \\
		\midrule
		Class/Model& CutPaste & Ours & CutPaste & Ours & CutPaste & Ours & SPADE & PaDiM & Ours \\
		\midrule
		Bottle     & 97.6 & \textbf{98.64}        & 98.3 & 100.00         & 100.0& 100.0                   & (98.4, 95.5)          & (98.3, 94.8) & (\textbf{98.98}, \textbf{96.80}) \\
		Cable      & 90.0 & \textbf{96.75}        & 80.6 & \textbf{97.62} & 96.2 & 97.59                   & (97.2, 90.9)          & (96.7, 88.8) & (\textbf{97.64}, \textbf{93.53}) \\
		Capsule    & 97.4 & \textbf{98.62}        & 96.2 & 93.15          & 95.4 & \textbf{97.68}          & (\textbf{99.0}, \textbf{93.7}) & (98.5, 93.5) & (98.98, 93.40) \\
		Carpet     & 98.3 & \textbf{99.29}        & 93.1 & 98.20          & \textbf{100.0}& 98.73          & (97.5, 94.7)          & (99.1, 96.2) & (\textbf{99.25}, \textbf{97.70}) \\
		Grid       & 97.5 & \textbf{98.53}        & \textbf{99.9} & 98.97 & 99.1 & 99.60                   & (93.7, 86.7)          & (97.3, 94.6) & (\textbf{98.99}, \textbf{96.08}) \\
		Hazelnut   & 97.3 & \textbf{98.81}        & 97.3 & 99.91          & 99.9 & \textbf{99.98}          & (\textbf{99.1}, 95.4) & (98.2, 92.6) & (98.89, \textbf{96.68}) \\
		Leather    & 99.5 & \textbf{99.51}        & 100.0& 100.00         & 100.0& 100.0                   & (97.6, 97.2)          & (98.9, 88.8) & (\textbf{99.66}, \textbf{99.35}) \\
		Metal Nut  & 93.1 & \textbf{97.59}        & \textbf{99.3} & 98.45 & 98.6 & 99.26                   & (98.1, \textbf{94.4)} & (97.2, 85.6) & (\textbf{98.56}, 91.65) \\
		Pill       & 95.7 & \textbf{98.34}        & 92.4 & 93.02          & 93.3 & \textbf{96.82}          & (96.5, 94.6)          & (95.7, 92.7) & (\textbf{98.95}, \textbf{95.39}) \\
		Screw      & 96.7 & \textbf{98.40}        & 86.3 & 85.94          & 86.6 & \textbf{91.89}          & (\textbf{98.9}, \textbf{96.0}) & (98.5, 94.4) & (98.86, 95.30) \\
		Tile       & 90.5 & \textbf{95.80}        & 93.4 & 98.40          & 99.8 & \textbf{99.88}          & (87.4, 75.9)          & (94.1, 86.0) & (\textbf{98.01}, \textbf{94.34}) \\
		Toothbrush & 98.1 & \textbf{99.00}        & 98.3 & \textbf{99.86} & 90.7 & 99.65                   & (97.9, 93.5)          & (98.8, 93.1) & (\textbf{98.93}, \textbf{95.06}) \\
		Transistor & 93.0 & \textbf{97.69}        & 95.5 & 93.04          & \textbf{97.5} & 95.21          & (94.1, \textbf{87.4)} & (97.5, 84.5) & (\textbf{97.99}, 81.40) \\
		Wood       & \textbf{95.5} & 95.00        & 98.6 & 98.59          & \textbf{99.8} & 99.12          & (88.5, \textbf{97.4)} & (94.9, 91.1) & (\textbf{96.65}, 95.79) \\
		Zipper     & \textbf{99.3} & 98.98        & 99.4 & 96.15          & \textbf{99.9} & 98.48          & (96.5, 92.6)          & (98.5, 95.9) & (\textbf{99.08}, \textbf{96.60}) \\
		\midrule
		Average    & 96.0 & \textbf{98.06}        & 95.2 & 96.75          & 97.1 & \textbf{98.26}          & (96.0, 91.7) & (97.5, 92.1) & (\textbf{98.62}, \textbf{94.60}) \\
		\bottomrule
	\end{tabular}
\end{table*}

\subsection{Experimental setup}
\label{subsec:setup_eval}
We conduct unsupervised anomaly detection (image-level) and localization (pixel-level segmentation) experiments using the MVTec~\cite{Bergmann_2019_CVPR} dataset with factory defects and the STC~\cite{Luo_2017_ICCV} dataset with surveillance camera videos. The code is in PyTorch~\cite{paszke2017automatic} with the FrEIA library~\cite{ardizzone2018analyzing} used for generative normalizing flow modeling.

Industrial MVTec dataset comprises 15 classes with total of 3,629 images for training and 1,725 images for testing. The train dataset contains only anomaly-free images without any defects. The test dataset contains both images containing various types of defects and defect-free images. Five classes contain different types of textures (carpet, grid, leather, tile, wood), while the remaining 10 classes represent various types of objects. We resize MVTec images without cropping according to the specified image resolution (\eg $H \times W = 256\times256, 512\times512$ \etc) and apply $\pm 5^{\circ}$ augmentation rotations during training phase only.

STC dataset contains 274,515 training and 42,883 testing frames extracted from surveillance camera videos and divided into 13 distinct university campus scenes. Because STC is significantly larger than MVTec, we experiment only with $256\times256$ resolution and apply the same pre-processing and augmentation pipeline as for MVTec.

We compare CFLOW-AD with the models reviewed in Section~\ref{sec:related} using MVTec and STC datasets. We use widely-used threshold-agnostic evaluation metrics for localization: area under the receiver operating characteristic curve (AUROC) and area under the per-region-overlap curve (AUPRO)~\cite{Bergmann_2019_CVPR}. AUROC is skewed towards large-area anomalies, while AUPRO metric ensures that both large and small anomalies are equally important in localization. Image-level AD detection is reported by the AUROC only.

We run each CFLOW-AD experiment four times on the MVTec and report mean ($\mu$) of the evaluation metric and, if specified, its standard deviation ($\pm \sigma$). For the larger STC dataset, we conduct only a single experiment. As in other methods, we train a separate CFLOW-AD model for each MVTec class and each STC scene. All our models use the same training hyperparameters: Adam optimizer with 2e-4 learning rate, 100 train epochs, 32 mini-batch size for encoder and cosine learning rate annealing with 2 warm-up epochs. Since our decoders are agnostic to feature map dimensions and have low memory requirements, we train and test CFLOW-AD decoders with 8,192 (32$\times$256) mini-batch size for feature vector processing. During the train phase 8,192 feature vectors are randomly sampled from 32 random feature maps. Similarly, 8,192 feature vectors are sequentially sampled during the test phase. The feature pyramid pooling setup for ResNet-18 and WideResnet-50 encoder is identical to PaDiM~\cite{defard2020padim}. The effects of other architectural hyperparameters are studied in the ablation study.

\subsection{Ablation study}
\label{subsec:ablation_eval}
Table~\ref{tab:ablation-results} presents a comprehensive study of various design choices for CFLOW-AD on the MVTec dataset using AUROC metric. In particular, we experiment with the input image resolution ($H \times W$), encoder architecture (ResNet-18~\cite{he}, WideResnet-50~\cite{BMVC2016_87}, MobileNetV3L~\cite{Howard_2019_ICCV}), type of normalizing flow (unconditional (UFLOW) or conditional (CFLOW)), number of flow coupling layers (\# of CL) and pooling layers (\# of PL).

Our study shows that the increase in number of decoder's coupling layers from 4 to 8 gives on average 0.15\% gain due to a more accurate distribution modeling. Even higher 1.4\% AUROC improvement is achieved when processing 3-scale feature maps (layers 1, 2 and 3) compared 2-scale only (layers 2, 3). The additional feature map (layer 1) with larger scale ($ H_1 \times W_1 = 32 \times 32$) provides more precise spatial semantic information. The conditional normalizing flow (CFLOW) is on average 0.5\% better than the unconditional (UFLOW) due to effective encoding of spatial prior. Finally, larger WideResnet-50 outperforms smaller ResNet-18 by 0.81\%. MobileNetV3L, however, could be a good design choice for both fast inference and high AUROC.

Importantly, we find that the optimal input resolution is not consistent among MVTec classes. The classes with macro objects \eg cable or pill tend to benefit from the smaller-scale processing (256$\times$256), which, effectively, translates to larger CNN receptive fields. Majority of classes perform better with 512$\times$512 inputs \ie smaller receptive fields. Finally, we discover that the transistor class has even higher AUROC with the resized to 128$\times$128 images. Hence, we report results with the highest performing input resolution settings in the Section~\ref{subsec:quant_eval} comparisons.

\begin{table}[t]
	\caption{Average AUROC and AUPRO on the MVTec~\cite{Bergmann_2019_CVPR} dataset, \%. Both the best detection and localization metrics are presented, if available. CFLOW-AD is with WideResNet-50 encoder.}
	\label{tab:mvtec-results}
	\centering
	\begin{tabular}{c|c|c|c}
		\toprule
		Metric & \multicolumn{2}{c|}{AUROC} & AUPRO\\
		\midrule
		Model & Detection & \multicolumn{2}{c}{Localization}\\
		\midrule
		DifferNet~\cite{rudolph2020differnet} &94.9&-&-\\
		DFR~\cite{DFR2020} &-&95.0&91.0\\
		SVDD~\cite{Yi_2020_ACCV} &92.1&95.7&-\\
		SPADE~\cite{cohen2021subimage} &85.5&96.0&91.7\\
		CutPaste~\cite{li2021cutpaste} &97.1&96.0&-\\
		PaDiM~\cite{defard2020padim} &97.9&97.5&92.1\\
		\midrule
		CFLOW-AD (ours) &\textbf{98.26}&\textbf{98.62}&\textbf{94.60} \\
		\bottomrule
	\end{tabular}
\end{table}

\begin{table}[t]
	\caption{Average AUROC on the STC~\cite{Luo_2017_ICCV} dataset, \%. Both the best available detection and localization metrics are showed. CFLOW-AD is with WideResNet-50 encoder.}
	\label{tab:stc-results}
	\centering
	\begin{tabular}{c|c|c}
		\toprule
		Metric & \multicolumn{2}{c}{AUROC}\\
		\midrule
		Model & Detection & Localization\\
		\midrule
		CAVGA~\cite{cavga} &-&85.0\\
		SPADE~\cite{cohen2021subimage} &71.9&89.9\\
		PaDiM~\cite{defard2020padim} &-&91.2\\
		\midrule
		CFLOW-AD (ours) &\textbf{72.63}&\textbf{94.48}\\
		\bottomrule
	\end{tabular}
\end{table}

\subsection{Quantitative comparison}
\label{subsec:quant_eval}
Table~\ref{tab:mvtec-results} summarizes average MVTec results for the best published models. CFLOW-AD with WideResNet-50 encoder outperforms state-of-the-art by 0.36\% AUROC in detection, by 1.12\% AUROC and 2.5\% AUPRO in localization, respectively. Table~\ref{tab:class-results} contains per-class comparison for the subset of models grouped by the task and type of encoder architecture. CFLOW-AD is on par or significantly exceeds the best models in per-class comparison with the same encoder setups.

Table~\ref{tab:stc-results} presents high-level comparison of the best recently published models on the STC dataset. CFLOW-AD outperforms state-of-the-art SPADE~\cite{cohen2021subimage} by 0.73\% AUROC in anomaly detection and PaDiM~\cite{defard2020padim} by 3.28\% AUROC in anomaly localization tasks, respectively.

Note that our CFLOW-AD models in Tables~\ref{tab:class-results}-\ref{tab:mvtec-results} use variable input resolution as discussed in the ablation study: 512$\times$512, 256$\times$256 or 128$\times$128 depending on the MVTec class. We used fixed 256$\times$256 input resolution in Table~\ref{tab:stc-results} for the large STC dataset to decrease training time. Other reference hyperparameters in Tables~\ref{tab:mvtec-results}-\ref{tab:stc-results} are set as: WideResnet-50 encoder with 3-scale pooling layers, conditional normalizing flow decoders with 8 coupling layers.

\begin{figure*}[ht]
	\centering
	\includegraphics[width=0.98\textwidth]{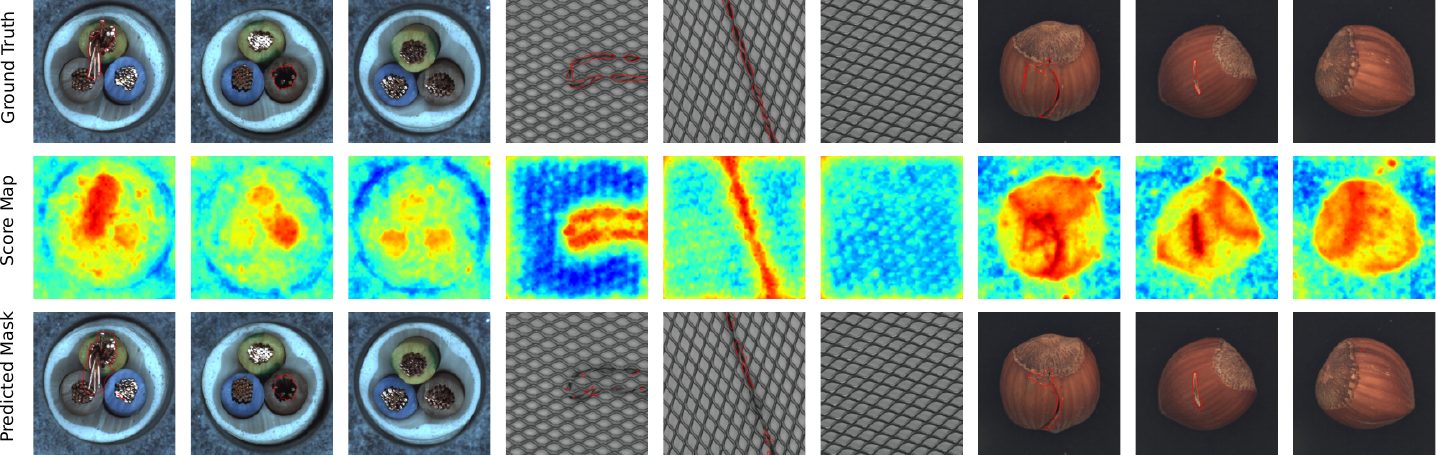}
	\caption{Examples of the input images with ground truth anomaly masks (top row) for various classes of the MVTec. Our CFLOW-AD model from Table~\ref{tab:mvtec-results} estimates anomaly score maps (middle row) and generates segmentation masks (bottom row) for a threshold selected to maximize F1-score. The predicted segmentation mask should match the corresponding ground truth as close as possible.}
	\label{fig:qual}
\end{figure*}

\begin{figure}[h]
	\centering
	\includegraphics[width=0.65\columnwidth]{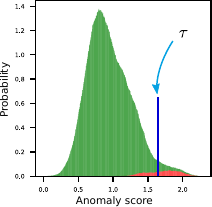}
	\caption{Distribution of anomaly scores for the cable class from the MVTec learned by CFLOW-AD model from Table~\ref{tab:mvtec-results}. Green density represent scores for the anomaly-free feature vectors, while region of red-color density shows scores for feature vectors with anomalies. The threshold $\tau$ is selected to optimize F1-score.}
	\label{fig:hist}
\end{figure}

\subsection{Qualitative results}
\label{subsec:qual_eval}
Figure~\ref{fig:qual} visually shows examples from the MVTec and the corresponding CFLOW-AD predictions. The top row shows ground truth masks from $\test$ including examples with and without anomalies. Then, our model produces anomaly score maps (middle row) using the architecture from Figure~\ref{fig:cflow}. Finally, we show the predicted segmentation masks with the threshold selected to maximize F1-score.

Figure~\ref{fig:hist} presents an additional evidence that our CFLOW-AD model actually addresses the OOD task sketched in Figure~\ref{fig:problem} toy example. We plot distribution of output anomaly scores for anomaly-free (green) and anomalous feature vectors (red). Then, CFLOW-AD is able to distinguish in-distribution and out-of-distribution feature vectors and separate them using a scalar threshold $\tau$.

\subsection{Complexity evaluations}
\label{sec:comp_eval}

\begin{table}[t]
	\caption{Complexity comparison in terms of inference speed (fps) and model size (MB). Inference speed for CFLOW-AD models from Table~\ref{tab:class-results} is measured for (256$\times$256) / (512$\times$512) inputs.}
	\label{tab:comp_eval}
	\centering
	\begin{tabular}{c|c|cc}
		\toprule
		\multirow{2}{*}{\shortstack{Complexity metric\\ and Model}}& \multirow{2}{*}{\shortstack{Inference\\speed, fps}}  & \multicolumn{2}{|c}{Model size, MB}\\
		& & STC & MVTec\\
		\midrule
		R18 encoder only      & 80 / 62       & \multicolumn{2}{c}{45}\\
		PaDiM-R18~\cite{defard2020padim}& 4.4 & 210 & 170\\
		CFLOW-AD-R18          & 34 / 12       & \multicolumn{2}{c}{96}\\
		\midrule
		WRN50 encoder only     & 62 / 30       & \multicolumn{2}{c}{268}\\
		SPADE-WRN50~\cite{cohen2021subimage}   & 0.1 & 37,000  & 1,400\\
		PaDiM-WRN50~\cite{defard2020padim}     & 1.1 & 5,200 & 3,800\\
		CFLOW-AD-WRN50         & 27 / 9        & \multicolumn{2}{c}{947}\\
		\midrule
		MNetV3 encoder only   & 82 / 61       & \multicolumn{2}{c}{12}\\
		CFLOW-AD-MNetV3       & 35 / 12       & \multicolumn{2}{c}{25}\\
		\bottomrule
	\end{tabular}
\end{table}

In addition to analytical estimates in Table~\ref{tab:comp_est}, we present the actual complexity evaluations for the trained models using inference speed and model size metrics. Particularly, Table~\ref{tab:comp_est} compares CFLOW-AD with the models from Tables~\ref{tab:mvtec-results}-\ref{tab:stc-results} that have been studied by Defard~\etal~\cite{defard2020padim}.

The model size in Table~\ref{tab:comp_eval} is measured as the size of all floating-point parameters in the corresponding model \ie its encoder and decoder (post-processing) models. Because the encoder architectures are identical, only the post-processing models are different. Since CFLOW-AD decoders do not explicitly depend on the feature map dimensions (only on feature vector depths), our model is significantly smaller than SPADE and PaDiM. If we exclude the encoder parameters for fair comparison, CFLOW-AD is 1.7$\times$ to 50$\times$ smaller than SPADE and 2$\times$ to 7$\times$ smaller than PaDiM.

Inference speed in Table~\ref{tab:comp_eval} is measured with INTEL I7 CPU for SPADE and PaDiM in Defard~\etal~\cite{defard2020padim} study with 256$\times$256 inputs. We deduce that this suboptimal CPU choice was made due to large memory requirements for these models in Table~\ref{tab:comp_eval}. Thus, their GPU allocation for fast inference is infeasible. In contrast, our CFLOW-AD can be processed in real-time with 8$\times$ to 25$\times$ faster inference speed on 1080 8GB GPU with the same input resolution and feature extractor. In addition, MobileNetV3L encoder provides a good trade-off between accuracy, model size and inference speed for practical inspection systems.

\section{Conclusions}
\label{sec:conclusion}
We proposed to use conditional normalizing flow framework to estimate the exact data likelihoods which is infeasible in other generative models. Moreover, we analytically showed the relationship of this framework to previous distance-based models with multivariate Gaussian prior.

We introduced CFLOW-AD model that addresses the complexity limitations of existing unsupervised AD models by employing fully-convolutional translation-equivariant architecture. As a result, CFLOW-AD is faster and smaller by a factor of 10$\times$ than prior models with the same input resolution and feature extractor setup.

CFLOW-AD achieves new state-of-the-art for popular MVTec with 98.26\% AUROC in detection, 98.62\% AUROC and 94.60\% AUPRO in localization. Our new state-of-the-art for STC dataset is 72.63\% and 94.48\% AUROC in detection and localization, respectively. Our ablation study analyzed design choices for practical real-time processing including feature extractor choice, multi-scale pyramid pooling setup and the flow model hyperparameters.


{\small
\bibliographystyle{ieee_fullname}
\bibliography{paper}
}

\ifwacvfinal
\appendix
\section{Relationship with the flow framework}
The loss function for the \textit{reverse} $D_{KL} \left[ \hat{p}_{Z} (\vz, \vtheta) \| p_{Z} (\vz) \right]$ objective~\cite{JMLR:v22:19-1028}, where $\hat{p}_{Z} (\vz, \vtheta)$ is the model prediction and $p_{Z} (\vz)$ is a target density, is defined as
\begin{equation} \label{eq:th5-1}
\mathcal{L}(\vtheta) = \mathbb{E}_{\hat{p}_{Z} (\vz, \vtheta)} \left[ \log \hat{p}_{Z} (\vz, \vtheta) - \log p_{Z} (\vz) \right]. \tag{5}
\end{equation}

The first term in~(5) can be written using~(4) definition for a standard MVG prior $(\vu \sim \mathcal{N}(\vzero, \mI))$ as
\begin{equation} \label{eq:th5-2}
\log \hat{p}_{Z} (\vz, \vtheta) = \log (2\pi)^{-D/2} - E^2(\vu)/2 + \log \left| \det \mJ \right|, \tag{5.1}
\end{equation}
where $E^2(\vu) = \| \vu \|_2^2$ is a squared Euclidean distance of $\vu$.

Similarly, the second term in~(5) can be written for MVG density~(2) using a square of Mahalanobis distance as
\begin{equation} \label{eq:th5-3}
\log p_{Z} (\vz) = \log (2\pi)^{-D/2} + \log \det \mSigma^{-1/2} - M^2(\vz)/2. \tag{5.2}
\end{equation}

By substituting~(5.1-5.2) into~(5), the constants $\log (2\pi)^{-D/2}$ are eliminated and the loss is
\begin{equation} \label{eq:supp_5-4}
\mathcal{L}(\vtheta) = \mathbb{E}_{\hat{p}_{Z} (\vz, \vtheta)} \left[ \frac{M^{2}(\vz) - E^{2}(\vu)}{2} + \log \frac{| \det \mJ |}{ \det \mSigma^{-1/2}} \right]. \tag{6}
\end{equation}

\section{CFLOW decoders for likelihood estimation}
We train CFLOW-AD using a maximum likelihood objective, which is equivalent to minimizing the \textit{forward} $D_{KL}$ objective~\cite{JMLR:v22:19-1028} with the loss defined by

\begin{equation} \label{eq:cflow1-1}
\mathcal{L}(\vtheta) = D_{KL} \left[ p_{Z} (\vz) \right \| \hat{p}_{Z} (\vz, \vc, \vtheta)] \tag{7},
\end{equation}
where $\hat{p}_{Z} (\vz, \vc, \vtheta)$ is a conditional normalizing flow (CFLOW) model with a condition vector $\vc \in\mathbb{R}^{C}$.

The target density $p_{Z} (\vz)$ is usually replaced by a constant because the parameters $\vtheta$ do not depend on this density during gradient-based optimization. Then by analogy with unconditional flow~(4), the loss~(7) for $\hat{p}_{Z} (\vz, \vc, \vtheta)$ can be written as
\begin{equation} \label{eq:cflow1-2}
\mathcal{L}(\vtheta) = -\mathbb{E}_{p_{Z}(\vz)} \left[ \log p_{U} (\vu) + \log \left| \det \mJ \right| \right] + \text{const}. \tag{7.1}
\end{equation}

In practice, the expectation operation in~(7.1) is replaced by an empirical train dataset $\train$ of size $N$. Using the definition of base distribution with $p_{U} (\vu)$, the final form of ~(7) can be expressed as
\begin{equation} \label{eq:cflow1-3}
\mathcal{L}(\vtheta) \approx \frac{1}{N} \sum^N_{i=1} \left[ \frac{\| \vu_i \|_2^2}{2} - \log \left| \det \mJ_i \right| \right] + \text{const}, \tag{7.2}
\end{equation}
where the random variable $\vu_i = g^{-1}(\vz_i, \vc_i, \vtheta)$ and the Jacobian $\mJ_i = \nabla_{\vz} g^{-1}(\vz_i, \vc_i, \vtheta)$ depend both on input features $\vz_i$ and conditional vector $\vc_i$ for CFLOW model.
\fi

\end{document}